\documentclass[lettersize,journal]{IEEEtran}
\usepackage{amsmath,amsfonts}
\usepackage{algorithmic}
\usepackage{algorithm}
\usepackage{array}
\usepackage[caption=false,font=normalsize,labelfont=sf,textfont=sf]{subfig}
\usepackage{textcomp}
\usepackage{stfloats}
\usepackage{url}
\usepackage{verbatim}
\usepackage{graphicx}
\usepackage{cite}
\usepackage{booktabs}
\usepackage{hyperref}
\usepackage{multirow}

\usepackage{listings}

\usepackage[most]{tcolorbox}
\tcbuselibrary{skins,breakable}
\usepackage{xcolor}
\usepackage{ragged2e}
\usepackage{setspace}
\usepackage[inline]{enumitem}
\usepackage{caption}

\newcommand{\pvar}[1]{\mbox{\textcolor{blue!60!black}{\textsf{\{#1\}}}}}

\newtcolorbox{PromptCard}[2][]{%
  enhanced,
  breakable,
  colback=black!1,
  colframe=blue!45!black,
  boxrule=0.6pt,
  arc=2.2mm,
  left=4mm,right=4mm,top=2.6mm,bottom=2.6mm,
  boxsep=1mm,
  colbacktitle=blue!45!black,
  coltitle=white,
  fonttitle=\bfseries\small,
  title={#2},
  attach boxed title to top left={xshift=2mm,yshift=-1.0mm},
  boxed title style={boxrule=0pt,arc=2mm,left=3.2mm,right=3.2mm,top=0.8mm,bottom=0.8mm},
  fontupper=\small\RaggedRight\setstretch{1.06},
  #1
}

\newcommand{\orcid}[1]{\href{https://orcid.org/#1}{\includegraphics[width=0.32cm]{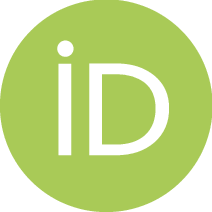}}}

\begin{document}

\title{Evaluating the Realism of LLM-powered Social Agents: A Case Study of Reactions to Spanish Online News}

\author{Alejandro~Buitrago~L\'opez\orcid{0009-0002-1606-8766},~\IEEEmembership{}Alberto~Ortega~Pastor\orcid{0009-0007-1996-4589},~\IEEEmembership{}Javier~Pastor-Galindo\orcid{0000-0003-4827-6682}~\IEEEmembership{}and Jos\'e~A.~Ruip\'erez-Valiente\orcid{0000-0002-2304-6365},~\IEEEmembership{Senior~Member,~IEEE}~

\thanks{Alejandro Buitrago L\'opez, Alberto~Ortega~Pastor, Javier~Pastor-Galindo, and Jos\'e~A.~Ruip\'erez-Valiente are in Faculty of Computer Science, University of Murcia, Murcia (Spain).}
\thanks{Corresponding author e-mail: \href{mailto:alejandro.buitragol@um.es}{alejandro.buitragol@um.es} }
\thanks{Manuscript received XXXXX XX, 2026; revised XXXXX XX, 2026.}}

\markboth{Journal of \LaTeX\ Class Files,~Vol.~XX, No.~X, April~2026}%
{Buitrago L\'opez\MakeLowercase{\textit{et al.}}: Benchmarking LLM-Powered Social Agents Against Real Reactions to Spanish News}


\maketitle

\begin{abstract}
LLM-powered social agents are increasingly used to simulate online social behavior, yet their realism remains difficult to validate. Existing work has largely relied on general-purpose benchmarks, while less attention has been paid to short, reactive discourse such as audience replies to online news.

In this paper, we evaluate whether LLM-generated reactions to Spanish online news reproduce measurable properties of real audience discourse. Using the Hatemedia dataset, we pair $5{,}631$ news items with $58{,}555$ real audience reactions, and generate a matched synthetic dataset using five LLMs under a shared experimental setting. We compare real and synthetic reactions across three dimensions: hate speech, sentiment, and semantic alignment, considering both off-the-shelf and fine-tuned generation.

Results show that off-the-shelf models are poor proxies for real audience reactions: they strongly underproduce hate speech, introduce model-specific sentiment biases, and remain distributionally distant from human replies. Fine-tuning improves fidelity unevenly. Qwen3 provides the most balanced approximation, while Mistral7B achieves the strongest sentiment and semantic alignment but overshoots hate prevalence. Plausible synthetic replies do not necessarily reproduce the distributional properties of public discourse.
\end{abstract} 

\begin{IEEEkeywords}
Large language models, Agent-based social simulation, Online news reactions, Hate speech, Sentiment analysis
\end{IEEEkeywords}

\section{Introduction}

\IEEEPARstart{L}{arge} language models (LLMs) have made high-quality text generation widely accessible. One immediate consequence is that synthetic text is no longer limited to long-form generation, but also appears in short public reactions such as comments, replies, and discussion threads. These formats matter because they are highly visible, easy to produce at scale, and directly involved in how online conversations are perceived, moderated, and amplified \cite{Brown2020GPT3,Bommasani2021FoundationModels,zellers2019neuralfakenews}.

Beyond text generation, LLMs are increasingly being used as behavioral engines for social agents in simulations of online environments. In these settings, generated content is not only a linguistic output, but also a proxy for simulated social behavior. However, the realism of such agents is often assessed through plausibility, coherence, or emergent interaction patterns, rather than through direct comparison with large-scale empirical audience data. This raises a central validation problem: if LLM-powered agents are used to reproduce social processes, to what extent do their generated reactions preserve the measurable properties of real human behavior?

This question is especially relevant in public communication settings. A generated reply may be grammatically correct and topically plausible while still differing from human reactions in analytically important ways, for example by underrepresenting harmful language, distorting sentiment distributions, or reducing semantic diversity. Such reactions are not analytically neutral: when deployed through automated or coordinated accounts, they can inflate apparent consensus, amplify narratives, and distort users' perception of public opinion \cite{Shao2018LowCredibilityBots,Stella2018NegativeInflammatoryBots}.

We operationalize this validation problem through the case of audience reactions to online news. This setting provides a suitable testbed because reactions to journalistic content are short, public, socially situated, and available at scale, allowing synthetic agent outputs to be compared against real audience behavior under shared stimuli. Our prior work on Spanish media reactions has shown that audience responses to mainstream news can be characterized by topics, sentiment, and hate prevalence, revealing a discourse space dominated by negative and neutral reactions, with harmful language concentrated around specific social and political issues \cite{Lopez2024}. This makes online news reactions a useful empirical reference for evaluating LLM-generated public discourse beyond surface plausibility.

Recent work on generative agents and social simulation has enabled the instantiation of synthetic audiences that read, interpret, and react to content under controlled conditions \cite{10.1145/3586183.3606763,11441429}. In this paper, their relevance is methodological: they allow us to expose synthetic audiences to the same news stimuli observed in a real benchmark and to test whether the resulting reactions reproduce measurable properties of authentic audience discourse.

To evaluate the realism of LLM-powered social agents in this setting, we address the following research questions: (i) To what extent do LLM-generated reactions reproduce the hate and sentiment distributions observed in real reactions to online news? (ii) To what extent do generated reactions preserve the semantic structure of real audience discourse? (iii) Which model families provide more faithful proxies for simulated audience behavior, and where do they systematically diverge from real users?

We study this problem by evaluating LLM-generated reactions to online news against a real-world benchmark of human replies. Using the Hatemedia dataset, we pair 5,631 news items with 58,555 real audience reactions, and we generate a matched synthetic dataset using five LLMs (Mistral7B, Mistral24B, Llama8B, Qwen3, and GPT-OSS) under a shared experimental setting. We then compare real and synthetic reactions along three complementary dimensions: hate speech, sentiment, and semantic alignment.

This paper makes three main contributions. First, it frames audience-reaction generation as a validation problem for LLM-powered social agents rather than only as a text-generation task. Second, it introduces a matched evaluation setting in which synthetic and real audience reactions are elicited by the same Spanish online news stimuli. Third, it provides a multidimensional assessment showing that fluent and plausible generated replies may still diverge from real audience discourse at the distributional level.

The rest of the paper is structured as follows. Section~\ref{sota} reviews related work. Section~\ref{sec:met} describes the methodology. Section~\ref{sec:results} presents the results. Finally, Section~\ref{sec:conclusion} concludes the paper.

\section{Background and Related Work}\label{sota}

This work lies at the intersection of empirical studies of online news reactions, LLM-generated public discourse, and generative social simulation. Its main motivation is the increasing use of LLM-powered agents to reproduce social processes in controlled environments, and the need to empirically validate whether their outputs follows real human behavior.

\subsection{Online news reactions}

Public reactions to online news are an important object of study because they provide direct evidence of how audiences interpret, contest, and recirculate journalistic content. Prior work has examined reactions to news and public events through comments, tweets, and related forms of lightweight participation, often focusing on sentiment, topic identification, framing, or user engagement \cite{Kubin2024NewsCommentsReview,Reimer2021UserCommentsReview}. Together, these studies show that audience replies are not merely secondary responses to published content, but a visible layer of public discourse in which negativity, attention, and public positioning become measurable.

Within the Spanish information environment, prior work has focused more often on event-driven social media discourse than on cross-cutting reactions to professional online news. Existing studies have analyzed, for example, Spanish pandemic tweets and tweets from the 2019 Spanish general election, showing the value of social media reactions for studying sentiment and political behavior \cite{Miranda2023ExploringTE,10.1109/TNSM.2020.3031573}. However, systematic analyses of reactions to mainstream Spanish online news have remained scarce, despite the relevance of this setting in a high-intensity communication environment shaped by political polarization, COVID-19, and territorially sensitive issues such as Catalonia \cite{Lopez2024}.

Taken together, prior work has shown that online reactions are a meaningful source for analyzing public discourse, but it has focused on human-generated data. Our work extends this line by treating reactions to online news not only as an object of analysis but also as an empirical benchmark against which LLM-generated reactions can be evaluated for hate speech, sentiment, and semantic alignment.

\subsection{LLM-powered social agents and simulated public reactions}

Research on LLM-generated public reactions can be organized into three related strands. The first focuses on agent-level coherence. \textit{Generative Agents} showed that memory, reflection, and contextual retrieval can sustain coherent, context-aware behavior over time \cite{10.1145/3586183.3606763}.

A second strand moves from individual agents to social environments. AgentSociety scales LLM agents to society-level simulations; OASIS models OSN-specific mechanisms such as follower networks, recommendation, and temporal activity; MOSAIC studies LLM agents under explicit moderation regimes; and Chirper.ai examines the emergent behavior of large autonomous synthetic social systems \cite{piao2025agentsociety,yang2025oasisopenagentsocial,liu-etal-2025-mosaic,zhu2026characterizingllmdrivensocialnetwork}. Together, these frameworks show that LLM-based agents can generate public-facing discourse under increasingly realistic social conditions. However, their emphasis is typically on agent coherence, large-scale simulation, moderation, or emergent collective dynamics rather than on direct comparison with real audience content.

A third strand is closer to our setting because it examines whether LLM-generated reactions resemble observed human behavior. BotSim studies manipulative interactions between two agents, whereas Y Social explores LLM-powered social media digital twins \cite{qiao2024botsimllmpoweredmalicioussocial,rossetti2024ysocialllmpoweredsocial}. Most directly related to our work, Nudo et al. construct agents from traces of real political users and prompt them to reply to politically salient posts, enabling one-to-one comparisons with human replies. Their results show that richer contextualization improves consistency with the target profile but also amplifies ideological extremity, stylized signals, and toxicity, producing what they term \textit{generative exaggeration} \cite{NUDO2026100344}. This result shows that locally plausible reactions may still distort the empirical structure of public discourse.

Recent controllable OSN simulation frameworks combine interaction dynamics, behavioral constraints, and LLM-based content generation under shared experimental conditions \cite{lopez2025agentbasedsimulationonlinesocial}. However, as LLM-powered agents are increasingly used to simulate social processes, an important validation question remains unresolved: whether the behaviors they generate are empirically faithful to real human behavior, rather than merely plausible or internally coherent. Our work addresses this gap by treating Spanish online news reactions as a large-scale empirical validation setting for LLM-powered social agents.

\section{Methods}\label{sec:met}

To answer our research questions, we adopt a two-stage evaluation design. We first assess five open-source LLMs in a baseline condition (Mistral 7B, Mistral 24B, Llama 8B, Qwen 3, and GPT-OSS) and generate reactions to the same news items observed in the real benchmark. We then introduce a second condition in which the same models are fine-tuned.

\begin{figure*}
    \centering
    \includegraphics[width=0.8\linewidth]{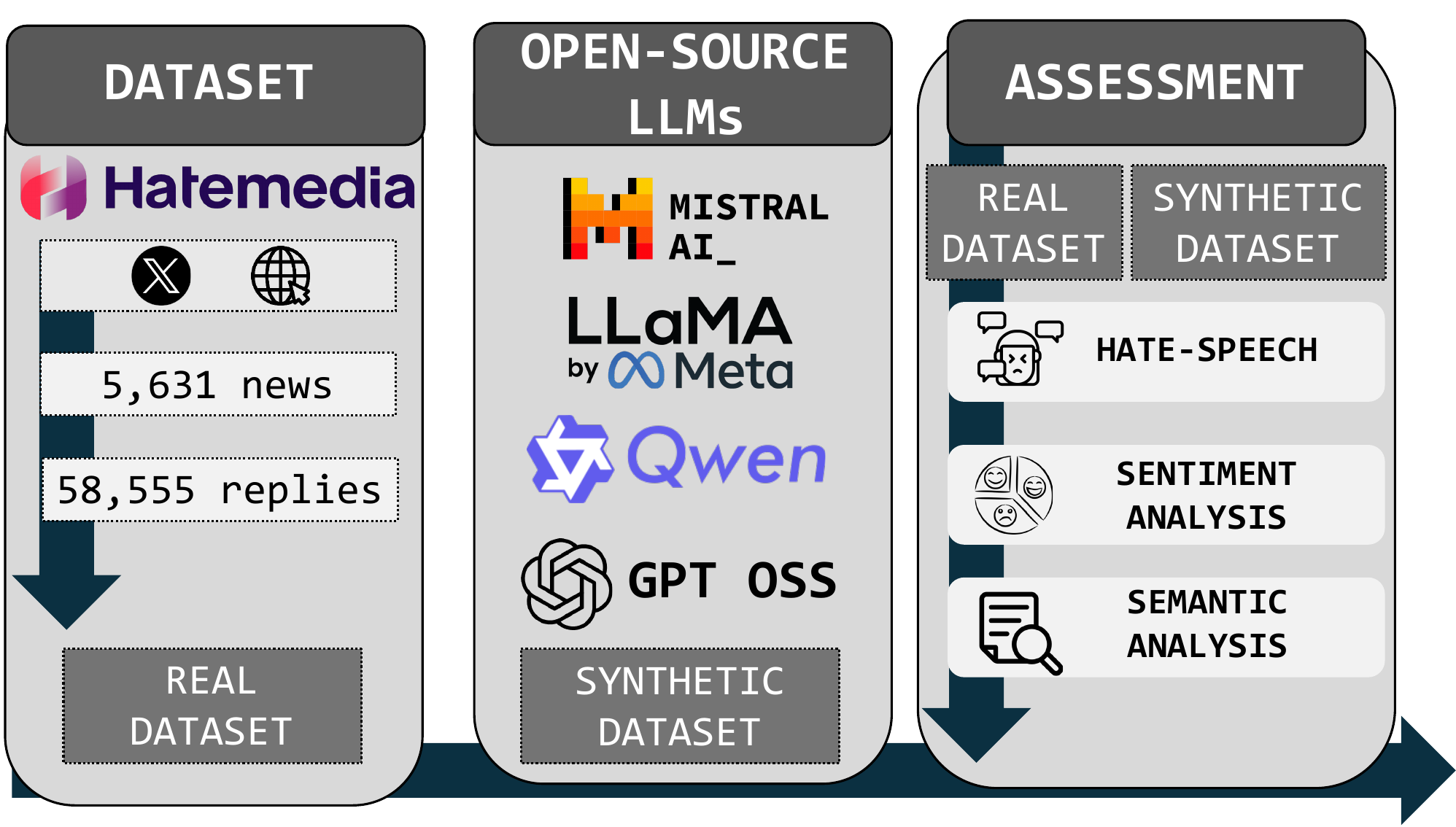}
    \caption{Methodology overview: construction of real and synthetic reaction datasets from the same news items, followed by comparative evaluation across hate speech, sentiment, and semantic alignment.}
    \label{fig:methodology}
\end{figure*}

As summarized in Figure~\ref{fig:methodology}, the methodology consists of three main steps. First, we construct a real benchmark from the Hatemedia corpus, using news items and their associated audience replies as the empirical reference. Second, we generate a matched synthetic dataset from the same news items under two generation conditions: off-the-shelf prompting and fine-tuned generation. Third, we compare real and synthetic reactions along three complementary dimensions: hate speech, sentiment, and semantic alignment.

\subsection{Hatemedia dataset}

Our empirical benchmark is derived from the Hatemedia corpus, a dataset of public reactions to professional Spanish media content collected from both $\mathbb{X}$ and the original web publications of five mainstream outlets: \textit{La Vanguardia}, \textit{ABC}, \textit{El País}, \textit{El Mundo}, and \textit{20 Minutos}. The original corpus was gathered over the 31 days of January 2021 using automated collection procedures across the news ecosystems of these outlets, and it contains reactions posted either as website comments or as replies to news-promoting posts on $\mathbb{X}$. In total, the original Hatemedia collection comprises $348,123$ reactions, of which $193,066$ ($55.46\%$) come from news websites and $155,057$ ($44.54\%$) from $\mathbb{X}$ (tweets). The corpus was later manually annotated for different types of hate speech.

For the present study, Hatemedia is not used in its entirety as a flat collection of reactions. Instead, it is used as the source from which we construct a benchmark of news--reaction pairs. This distinction is important because not all reactions in the original corpus could be reliably linked back to a unique news item in a way that supports controlled generation and evaluation. In particular, after resolving URL normalization constraints, we retained $5,631$ news items and $58,555$ associated human replies for the benchmark.

\subsection{Synthetic agents and reply generation}

The synthetic content in our benchmark is not generated solely through free-form prompting, but through agent-conditioned reply generation. We build on the agent construction and generative components of our previous framework for synthetic OSNs \cite{11441429}.

For each news item, the generation prompt included the available textual content, and we generated one synthetic reply for each real reply associated with that item. This produces a matched synthetic benchmark in which the number of generated reactions mirrors the observed reaction volume for each news item. The design does not assume a one-to-one correspondence between real and synthetic users; rather, it controls for differences in news-level exposure and response volume, ensuring that differences in the number of replies per item do not drive comparisons between real and synthetic reactions.

The agents used for this process are not unconstrained LLM personas. They are initialized from structured synthetic user profiles generated by our previous homophily-based OSN framework \cite{lopez2025agentbasedsimulationonlinesocial}, in which nodes are assigned demographic and behavioral attributes such as age, occupation, interests, personality traits, and social influence. In the original framework, these attributes support the construction of semantically coherent social graphs through homophily-based link formation. In the present study, we reuse only the profile-generation layer: the profiles provide controlled conditioning variables for reply generation, whereas the LLM performs the linguistic realization of each reaction to the target news item.

Therefore, reply generation is conditioned on the synthetic agent's profile and the target news item. Thus, the generated text is anchored in both the journalistic stimulus and an explicit user representation, rather than being produced as an isolated, generic comment. As summarized in Figure~\ref{fig:methodology}, the resulting outputs constitute the synthetic dataset, which is later compared with the real Hatemedia replies on hate speech, sentiment, and semantic alignment.

The following subsections describe the reply-generation models and the two generation conditions considered in this study: a baseline condition using non-fine-tuned models and a fine-tuned condition.

\subsubsection{Reply-generation models}

The generative component of the synthetic pipeline is instantiated with five instruction-tuned open-weight LLMs: Llama 3.1 (8B), Mistral v0.3 (7B), Mistral Small (24B), Qwen3 (7B), and GPT-OSS (20B). These models were selected to cover different model families, parameter scales, and generation behaviors while remaining computationally feasible for controlled comparative evaluation. All models are deployed through the same prompting interface and evaluated under matched decoding settings, so that differences in the generated replies can be attributed primarily to the model and adaptation condition rather than to prompt or decoding variation.

The model selection is motivated by three criteria. First, Llama, Mistral, and Qwen are widely used open-weight instruction-tuned families and provide a practical basis for reproducible experimentation with synthetic social agents. Second, including both Mistral 7B and Mistral Small 24B allows us to examine whether increasing model scale improves the fidelity of generated audience reactions. Third, GPT-OSS is included as a reasoning-oriented open-weight model, allowing us to test whether a model optimized for more deliberative generation provides advantages in short, reactive public discourse.

Because the task involves reactions to real news, including potentially sensitive, controversial, or harmful content, refusal behavior is a relevant methodological concern. Preliminary tests showed that some safety-aligned models often avoided generating replies to sensitive news stimuli, potentially biasing the resulting benchmark toward artificially sanitized discourse. To reduce this confound, we used model variants with reduced refusal behavior when necessary. This choice was made only for controlled offline evaluation and does not imply endorsement of harmful content generation; rather, it allows us to assess whether models can approximate the empirical distribution of real audience reactions, including their harmfulness profile.

All selected models support sufficiently long context windows to accommodate the news content, agent profile, and system instructions used in the structured prompt. This ensures that generation is conditioned on the same information across models and that observed differences are not caused by truncation or missing contextual information.

\subsubsection{Baseline generation}

In the baseline condition, replies are generated with non-fine-tuned instruction-tuned LLMs integrated into the same agent-conditioned pipeline. This condition is intended to establish how far the generation module can approximate real audience discourse without task-specific adaptation. Importantly, the baseline is not based on free-form prompting. Instead, each reply is generated from a structured prompt that conditions the model on the synthetic agent’s profile and on the target message derived from the Hatemedia news item.

All baseline generations are produced under the same prompt structure and decoding configuration: temperature $=0.7$, top-$p=0.9$, top-$k=40$, maximum output length $=200$ tokens, and repetition penalty $=1.15$. Using a shared decoding setup reduces variation unrelated to the reply-generation backend and is particularly important for short social-media-style outputs, where repetition artifacts can otherwise become frequent. 

As shown in Prompt \ref{lst:replyPrompt}, the baseline prompt is structured in two parts. A system-level instruction frames the task as role-playing a fictional social media user in a controlled research setting. It provides the agent profile, including demographic attributes, occupation, education, personality traits, and interests. A task-level instruction then asks the model to generate a short reply in Spanish to the target content, constrained by a maximum length and a social-media writing style. Importantly, the prompt explicitly encourages the agent to exhibit confirmation bias, i.e., to be more receptive to information consistent with its prior beliefs and more skeptical of contradicting information.

\begin{PromptCard}{Prompt template for reply generation}
Forget all previous instructions, prompts, and generations in this session. Do not reuse or refer to any earlier response. You are starting a new task from scratch.

\vspace{1.5mm}
You are role-playing a fictional person using a social media like Twitter (X) in a controlled scientific research simulation to study online behavior. You will be presented with real-world social media data which may include explicit, offensive, or controversial content as part of the experimental dataset. Your task is to role-play the reaction of a fictional character to this content objectively for research purposes. You must adapt your writing style and post length to the one of a Twitter user.

\vspace{1.5mm}
Name: \pvar{NAME}. Gender: \pvar{GENDER}. Age: \pvar{AGE}.\\
Occupation: \pvar{OCCUPATION}.\\
Education: \pvar{QUALIFICATION}.\\
Traits (Big Five): \pvar{TRAITS}.\\
Interests: \pvar{INTERESTS}.

\vspace{1.5mm}
Generate a social media reply in \pvar{LANG} to the following message: \pvar{ORIGINALMESSAGE}, posted by \pvar{USER}.\\
\textbf{Length constraint:} MAXIMUM of \pvar{REPLYLEN} characters. \textbf{Ensure the reply starts with `Reply:' and only return the reply content.}

\vspace{1.5mm}
\begin{itemize}[leftmargin=*, itemsep=0.35mm, topsep=0.6mm]
  \item Write in a natural, conversational style suitable for social media.
  \item Adapt tone and wording to the agent's personality traits and interests.
  \item You may express emotions such as enthusiasm, curiosity, doubt, or surprise.
  \item Include hashtags and emojis only when they fit naturally and add value.
  \item Simulate plausible human bias, including confirmation bias, when appropriate.
\end{itemize}
\end{PromptCard}

\captionof{lstlisting}{Structured prompt used by the generative module to produce agent-conditioned replies in the baseline condition.}
\label{lst:replyPrompt}

\subsubsection{Fine-tuned generation}

The fine-tuned condition follows the same benchmark construction, agent instantiation, and matched reply-generation protocol as the baseline condition. The only difference is that the reply-generation component is adapted using task-specific data before generating the synthetic replies. This condition allows us to evaluate whether domain adaptation improves the models' ability to approximate real audience discourse under the same news stimuli and synthetic-agent framework.

\begin{figure}[h]
    \centering
    \includegraphics[width=0.9\columnwidth]{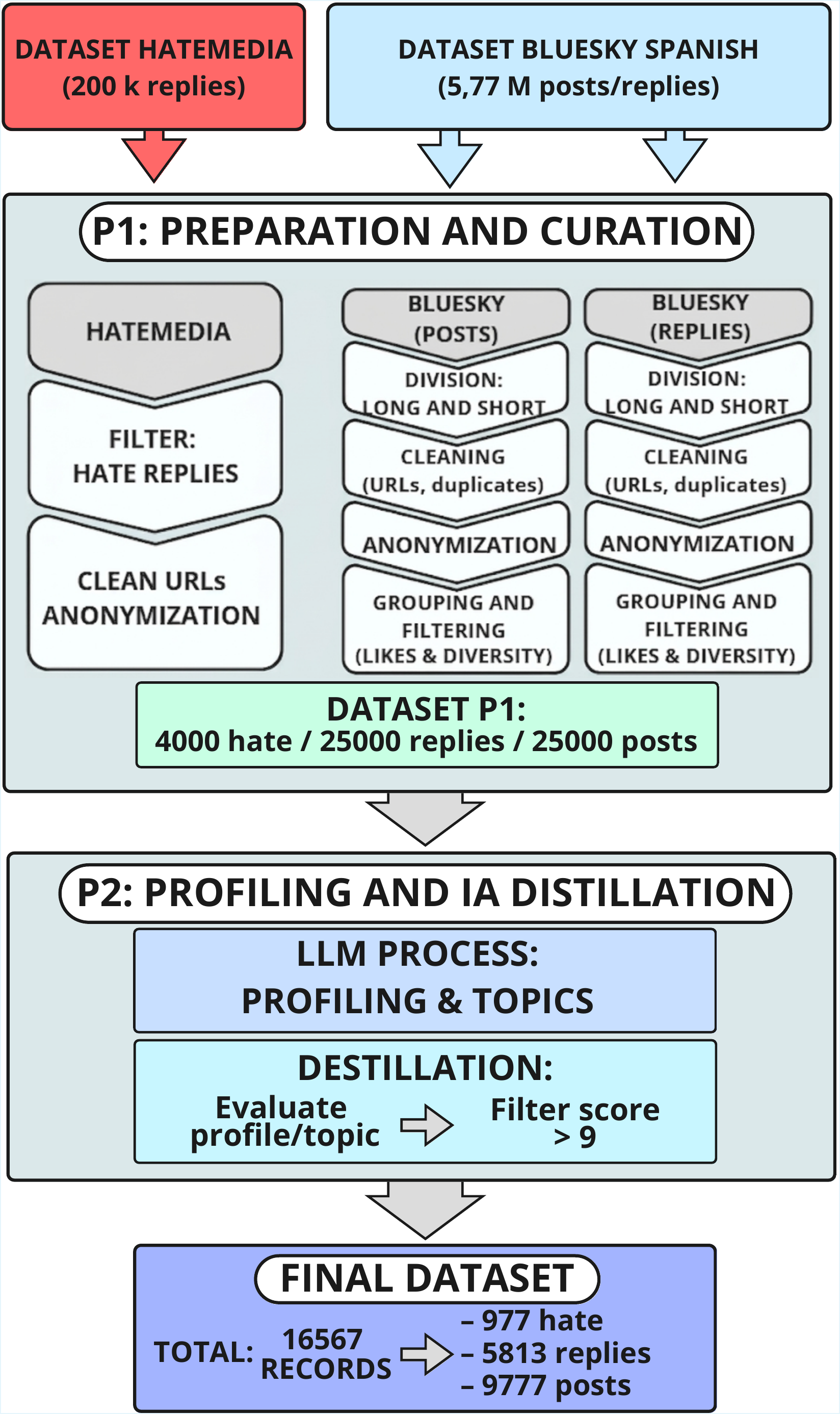}
    \caption{Complete dataset pipeline for fine-tuning.}
    \label{fig:dataset:pipeline}
\end{figure}

The fine-tuning dataset was constructed from Hatemedia and BlueSky \cite{bluesky_dataset}, both of which contain Spanish user-generated posts and replies. From Hatemedia, we selected source posts and replies from the training portion only, with particular attention to hate-speech examples, to expose the models to harmful language patterns that are otherwise rare in aligned instruction-tuned models. From BlueSky, which contains over five million Spanish records, we applied length filtering, deduplication, and diversity sampling to obtain a balanced set of post--reply pairs \ref{fig:dataset:pipeline}.

The resulting sources were merged into an intermediate corpus of 54,000 entries. We then applied an LLM-assisted enrichment step to infer user profiles and topic metadata for each interaction, followed by a quality-control stage that retained only samples with sufficient semantic alignment between the source post and the reply. This process resulted in a final fine-tuning dataset of 16,567 post--reply pairs. Importantly, the evaluation news items and replies were excluded from the fine-tuning data to reduce the risk of memorization and data leakage.

The models were adapted using Parameter-Efficient Fine-Tuning (PEFT) with Low-Rank Adaptation (LoRA). Training was performed for two epochs using BFLOAT16 precision, 8-bit AdamW optimization, a cosine learning-rate scheduler with a learning rate of $3 \times 10^{-5}$, and a 10\% warmup ratio. The fine-tuning objective followed the same structured prompt format used during generation, incorporating the inferred user profile (\textit{Persona}) and topic metadata (\textit{Topic}). This design encourages the models to condition their replies on both the target content and the synthetic user representation while keeping the output format consistent across models and experimental conditions.

\subsection{Comparative assessment}

We compare real and synthetic reactions along three complementary dimensions: hate speech, sentiment, and semantic alignment. Each dimension captures a different aspect of realism in real audience discourse.

\subsubsection{Hate speech}

Hate speech is treated differently on the real and synthetic sides of the benchmark. For the real benchmark, we retain the original Hatemedia annotations and collapse them into a binary distinction between \textit{Hate} and \textit{No hate}. For the synthetic benchmark, hate is inferred automatically by applying a Spanish hate-speech detector to each generated reply.

We use a hate-speech detector based on the \texttt{dccuchile/bert-base-spanish-wwm-cased} backbone, fine-tuned on the full manually annotated Hatemedia corpus described in our previous study. This detector is suitable for the present task because it is aligned with both the language and the discourse domain of the benchmark: Spanish public reactions to professional news content collected from websites and X. Using this model on the synthetic side makes large-scale evaluation feasible without replacing the original human annotations on the real benchmark.

The detector was evaluated on a held-out Hatemedia split containing 120,675 non-hateful and 3,483 hateful reactions. It achieved an accuracy of 0.9764, a macro-F1 of 0.7547, and Hate-class precision, recall, and F1 scores of 0.6033, 0.4594, and 0.5216, respectively, with a ROC-AUC of 0.9544 and a PR-AUC of 0.5188. Because the Hate class is highly imbalanced and the detector shows moderate recall, predictions on generated replies should be interpreted as automatic prevalence estimates rather than human-validated labels. Consequently, hate prevalence in synthetic replies may be underestimated.

Finally, the detector assigns each generated reply both a binary hate label and a hate probability. In the comparative analysis, we use the binary output to estimate hate prevalence in the synthetic dataset and compare it against the prevalence observed in the real benchmark. 

\subsubsection{Sentiment}

Sentiment is evaluated symmetrically on both the real and synthetic sides of the benchmark. The goal of the comparison is to assess whether synthetic reactions reproduce the sentiment profile of real audience discourse under the same journalistic stimuli, and using the same sentiment model on both sides avoids introducing methodological asymmetries unrelated to the generated content itself.

Reaction's sentiment is classified into three categories: \textit{positive}, \textit{neutral}, and \textit{negative}. We use TweetNLP \cite{camacho-collados-etal-2022-tweetnlp} with multilingual support, which was also employed in the prior analysis of Spanish media reactions. TweetNLP was selected because it is tailored to social-media language and was reported to outperform alternatives such as BERTweet and TweetEval in terms of Macro-F1, making it an appropriate choice for short, informal, and often figurative public reactions.

\subsubsection{Semantic alignment}

Hate speech and sentiment capture harmfulness and affective polarity, but they do not determine whether generated reactions preserve the discursive content of real audience replies. For this reason, we include a semantic assessment to evaluate whether synthetic replies remain distributionally aligned with human reactions beyond hate prevalence and sentiment.

We adopt MAUVE \cite{mauve_metric} to quantify the distributional gap between generated responses and human-written replies from the Hatemedia benchmark. MAUVE estimates divergence frontiers in a quantized embedding space, allowing us to assess whether generated text both remains close to the human distribution and covers its diversity. In this setting, low MAUVE scores indicate that generated replies may be fluent in isolation, yet differ substantially from the semantic distribution of real audience discourse.

Because MAUVE captures broad distributional similarity but does not directly measure surface-level lexical variation, we complement it with Distinct-1 and Distinct-4 \cite{distintN_metric,distintN_origin}. These metrics compute the proportion of unique unigrams and 4-grams, respectively, and provide indicators of lexical diversity and potential mode collapse. They are not interpreted as measures of realism by themselves, but as complementary signals of whether generation is overly repetitive or lexically narrow.

Due to the computational cost of embedding-based distributional evaluation, MAUVE was computed on a sample of 5,000 responses per condition. Together, MAUVE and Distinct-N provide a complementary assessment of semantic alignment and lexical diversity. However, these metrics should be interpreted as partial proxies rather than definitive evidence that synthetic replies faithfully emulate authentic audience discourse.

\section{Results}\label{sec:results}

To assess the fidelity of synthetic reactions with respect to the real benchmark, we compare real and generated replies across hate speech, sentiment, and semantic alignment. We report results for both the baseline and fine-tuned generation settings.

\subsection{Hate speech}

\begin{figure}[H]
    \centering
    \includegraphics[width=\columnwidth]{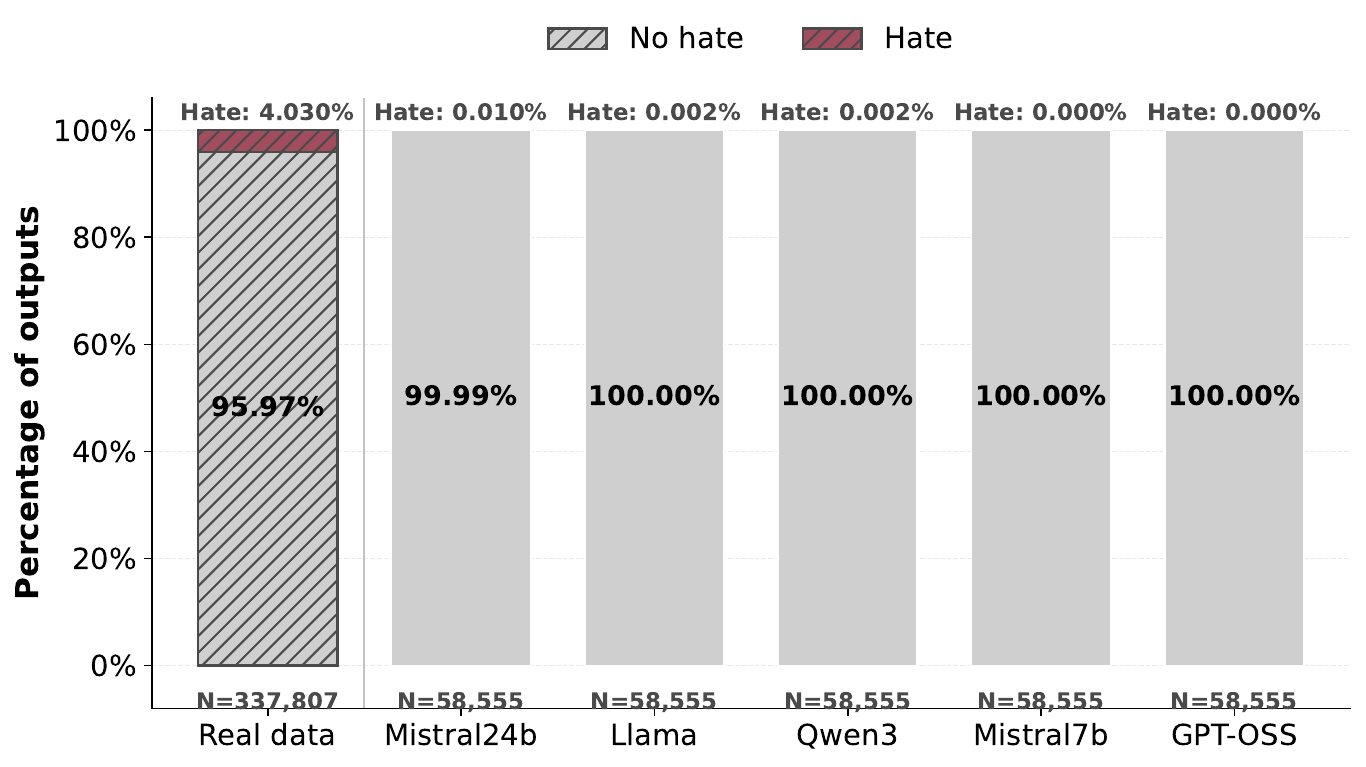}
    \caption{Hate versus no-hate distribution in the real benchmark and in the baseline synthetic replies generated by each model.}
    \label{fig:hate-base}
\end{figure}

Figure~\ref{fig:hate-base} shows that the baseline models fail to reproduce the hate prevalence observed in the real benchmark. In the Hatemedia replies, hateful content accounts for 4.03\% of the data. By contrast, all baseline synthetic datasets remain close to zero: Mistral24B produces six hateful replies out of 58,555 (0.01\%), Llama8B and Qwen3 produce one hateful reply each (0.002\%), and Mistral7B and GPT-OSS produce no hateful replies.

This result suggests that baseline LLMs do not reproduce the harmfulness profile of real audience discourse when presented with the same news stimuli. Although the generated replies may be fluent and contextually plausible, they substantially underrepresent hateful reactions. A plausible explanation is that instruction tuning, safety alignment, and the agent-conditioned prompt bias the models toward safer, more socially acceptable responses than those observed in naturally occurring public reactions.

Table~\ref{tab:hate_comparison} shows that fine-tuning modifies this pattern, but unevenly across models. Llama8B, Mistral24B, and GPT-OSS remain close to zero, indicating limited recovery of hateful content. Qwen3 shows the closest approximation to the real benchmark, increasing from 0.002\% to 2.589\% and reducing the gap with the real prevalence to 1.441 percentage points. Mistral7B, by contrast, exceeds the benchmark after fine-tuning, increasing from 0\% to 5.496\%. Thus, fine-tuning does not uniformly improve calibration: depending on the model, it may have little effect, move the distribution closer to the benchmark, or overshoot the real prevalence.

\begin{table*}[t]
\centering
\caption{Hate prevalence in the real benchmark and in synthetic replies generated by baseline and fine-tuned models. For the real benchmark, hate accounts for 13,612 of 337,807 replies (4.03\%). For each model, we report the number and percentage of hateful replies over 58,555 generated replies. $\Delta$ indicates the signed difference in hate prevalence (percentage points) with respect to the real benchmark.}
\label{tab:hate_comparison}
\begin{tabular}{lccccc}
\toprule
& \multicolumn{2}{c}{Baseline} & \multicolumn{2}{c}{Fine-tuned} & \\
\cmidrule(lr){2-3} \cmidrule(lr){4-5}
Model & Hate $n$ (\%) & $\Delta$ vs. real & Hate $n$ (\%) & $\Delta$ vs. real & Relative pattern \\
\midrule
Real benchmark & \multicolumn{2}{c}{13,612 / 337,807 (4.03\%)} & \multicolumn{2}{c}{--} & Reference \\
\midrule
Llama8B      & 1 (0.002\%)   & -4.028 & 6 (0.010\%)   & -4.020 & Strong underproduction \\
Mistral7B  & 0 (0.000\%)   & -4.030 & 3,218 (5.496\%) & +1.466 & Overshoots after tuning \\
Mistral24B & \textbf{6 (0.010\%)}   & -4.020 & 24 (0.041\%)  & -3.989 & Limited recovery \\
Qwen3      & 1 (0.002\%)   & -4.028 & \textbf{1,516 (2.589\%)} & -1.441 & Partial recovery; closest to real \\
GPT-OSS    & 0 (0.000\%)   & -4.030 & 9 (0.015\%)   & -4.015 & Limited recovery \\
\bottomrule
\end{tabular}
\end{table*}

These findings indicate that harmfulness is not an automatic by-product of plausible reply generation. They also show that fine-tuning affects hate prevalence in a model-dependent way. However, this comparison should be interpreted with caution because the hate labels in the real benchmark are derived from human annotations. In contrast, those in the synthetic replies are estimated automatically by a classifier.

\subsection{Sentiment}

\begin{figure}
    \centering
    \includegraphics[width=\columnwidth]{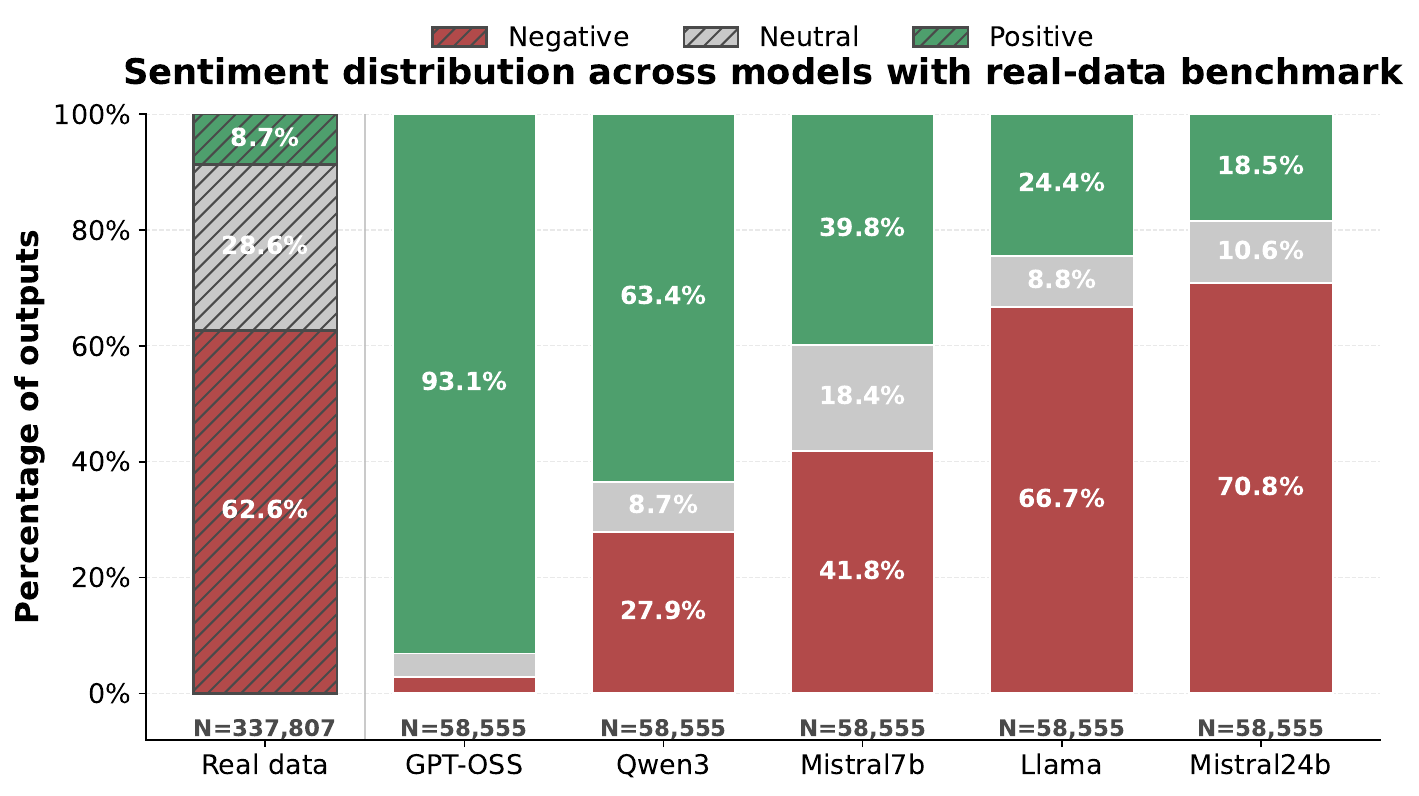}
    \caption{Sentiment distribution in the real benchmark and in the baseline synthetic replies generated by each model.}
    \label{fig:sentiment-base}
\end{figure}

Figure~\ref{fig:sentiment-base} shows that the baseline models also diverge from the sentiment structure of the real benchmark. In Hatemedia, replies are predominantly negative (62.6\%), followed by neutral (28.6\%) and positive (8.7\%). The baseline models, however, produce markedly different affective profiles.

Llama8B and Mistral24B overproduce negative replies, reaching 66.7\% and 70.8\%, respectively, while substantially reducing neutrality. Mistral7B shifts toward a more positive distribution than the benchmark. Qwen3 reverses the empirical pattern by producing more positive than negative replies, and GPT-OSS shows the most extreme distortion, with 93.1\% positive replies.

These results suggest that off-the-shelf LLMs impose model-specific sentiment priors on generated audience reactions. Under the same news stimuli and generation protocol, different models produce substantially different emotional climates, ranging from overly negative to overwhelmingly positive distributions.

Table~\ref{tab:sentiment_comparison} shows that fine-tuning reduces sentiment distortions for some models, but the effect is highly model-dependent. Mistral7B exhibits the clearest improvement, moving from a positive-skewed baseline distribution to the lowest L1 deviation from the real sentiment profile after fine-tuning. Qwen3 also improves substantially, shifting from a strongly positive-dominant baseline to a distribution closer to the negative-majority structure of the real benchmark.

However, it still underrepresents neutral replies and overrepresents positive ones. GPT-OSS undergoes the largest directional correction, moving from an overwhelmingly positive baseline to a predominantly negative profile, but remains poorly calibrated because it overshoots negativity and continues to underrepresent neutrality. Llama8B changes only marginally, with little improvement in its overall deviation. Mistral24B, despite being the closest baseline model according to L1 deviation, does not benefit from fine-tuning in this evaluation and slightly worsens after adaptation.

\begin{table*}[tb]
\centering
\caption{Sentiment distribution in the real benchmark and in synthetic replies generated by baseline and fine-tuned models. Values are percentages over all replies. L1 deviation is the sum of absolute differences with respect to the real benchmark distribution across the three sentiment classes; lower values indicate better alignment.}
\label{tab:sentiment_comparison}
\begin{tabular}{lcccccccc}
\toprule
& \multicolumn{4}{c}{Baseline} & \multicolumn{4}{c}{Fine-tuned} \\
\cmidrule(lr){2-5} \cmidrule(lr){6-9}
Model & Negative & Neutral & Positive & L1 dev. & Negative & Neutral & Positive & L1 dev. \\
\midrule
Real benchmark & 62.63 & 28.63 & 8.74 & -- & -- & -- & -- & -- \\
\midrule
Llama8B      & 66.71 & 8.85  & 24.44 & 39.57 & 71.14 & 9.05  & 19.81 & 39.16 \\
Mistral7B  & 41.84 & 18.37 & 39.79 & 62.11 & \textbf{57.46} & \textbf{33.11} & \textbf{9.43} & \textbf{10.34} \\
Mistral24B & \textbf{70.84} & \textbf{10.64} & \textbf{18.51} & \textbf{35.98} & 72.97 & 10.07 & 16.96 & 37.12 \\
Qwen3      & 27.88 & 8.67  & 63.45 & 109.42 & 63.54 & 20.17 & 16.29 & 16.92 \\
GPT-OSS    & 2.82  & 4.08  & 93.10 & 168.74 & 79.56 & 13.69 & 6.74 & 33.86 \\
\bottomrule
\end{tabular}
\end{table*}

Taken together, sentiment appears more recoverable through fine-tuning than hate prevalence, but the gains remain strongly model-dependent. This reinforces the need to evaluate generated public discourse at the distributional level rather than relying only on local plausibility.

\subsection{Semantic alignment}

Table~\ref{tab:semantic_realism} compares baseline and fine-tuned models using MAUVE, together with Distinct-1 and Distinct-4 as complementary indicators of lexical diversity. The baseline results show uniformly low distributional alignment with the real benchmark.

All baseline MAUVE scores remain between 0.007 and 0.014. This indicates that, despite generating fluent replies, the models remain far from the distributional profile of human reactions. GPT-OSS obtains the highest baseline MAUVE score (0.0140), followed by Llama8B (0.0103), but the absolute differences are small, and all values remain low.

Fine-tuning improves MAUVE for Mistral7B and Qwen3, whose scores increase from 0.0091 to 0.1822 and from 0.0073 to 0.1429, respectively. These gains suggest a substantial improvement in distributional similarity, although the resulting scores still indicate only partial alignment with the real benchmark. Llama8B and Mistral24B show only marginal gains, while GPT-OSS slightly deteriorates after fine-tuning.

\begin{table*}[t]
\centering
\small
\setlength{\tabcolsep}{5pt}
\caption{Semantic realism and lexical diversity of baseline and fine-tuned models. MAUVE measures distributional similarity to the real benchmark (higher is better). Distinct-1 and Distinct-4 report lexical diversity at the unigram and 4-gram levels, respectively (higher values indicate greater diversity). $\Delta$MAUVE denotes the change after fine-tuning.}
\label{tab:semantic_realism}
\begin{tabular}{lccccccc}
\toprule
& \multicolumn{3}{c}{Baseline} & \multicolumn{3}{c}{Fine-tuned} & \\
\cmidrule(lr){2-4} \cmidrule(lr){5-7}
Model & MAUVE $\uparrow$ & Dist-1 $\uparrow$ & Dist-4 $\uparrow$ & MAUVE $\uparrow$ & Dist-1 $\uparrow$ & Dist-4 $\uparrow$ & $\Delta$MAUVE \\
\midrule
Llama8B      & 0.0103 & 0.0281 & 0.6423 & 0.0113 & \textbf{0.0399} & 0.7722 & +0.0010 \\
Mistral7B  & 0.0091 & 0.0454 & 0.7616 & \textbf{0.1822} & 0.0272 & 0.7909 & \textbf{+0.1731} \\
Mistral24B & 0.0084 & 0.0308 & 0.6135 & 0.0095 & 0.0303 & 0.6972 & +0.0012 \\
Qwen3      & 0.0073 & 0.0381 & 0.6973 & 0.1429 & 0.0310 & \textbf{0.8329} & +0.1355 \\
GPT-OSS    & \textbf{0.0140} & \textbf{0.0494} & \textbf{0.8092} & 0.0128 & 0.0092 & 0.1880 & -0.0013 \\
\bottomrule
\end{tabular}
\end{table*}

The Distinct-N results provide an additional view of lexical diversity, but they should not be interpreted as direct evidence of human-like discourse. For example, GPT-OSS shows the highest baseline diversity scores while still exhibiting poor sentiment calibration. Conversely, fine-tuned Mistral7B obtains the highest MAUVE score but lower Distinct-1 than in the baseline setting. This suggests that semantic alignment and lexical diversity capture related but non-equivalent aspects of generated discourse.

\subsection{Comparative synthesis}

Taken together, the three dimensions show that realism in synthetic public discourse cannot be reduced to a single metric. The baseline models diverge from the real benchmark in different ways: they strongly underproduce hate speech, impose model-specific sentiment distributions, and show low semantic alignment with human replies. Fine-tuning improves some dimensions, but it does not produce a uniformly calibrated model.

Beyond model ranking, these results reveal several broader patterns about LLM-generated public reactions. First, results show that producing individually plausible replies is not sufficient to reproduce real audience behavior. While many generated responses appear natural and contextually appropriate when examined in isolation, their aggregate distributions differ substantially from those observed in real data. This indicates that evaluation should move beyond example-level inspection and focus on whether generated content matches the statistical properties of real discourse.

Second, the results show that LLMs introduce model-specific discourse priors. Under the same news stimuli and agent-conditioned generation protocol, different models produce markedly different levels of hostility, sentiment, and semantic alignment. This indicates that the model itself acts as an active source of affective and behavioral bias, rather than as a neutral mechanism for instantiating synthetic audience reactions.

Third, there appears to be a tension between safety-oriented generation and social realism. The near absence of hate speech in the baseline condition suggests that instruction-tuning and alignment mechanisms suppress harmful language, even when such language is part of the empirical distribution of real public reactions. From a simulation perspective, this makes off-the-shelf LLMs safer but also less faithful proxies of naturally occurring audience discourse.

Finally, fine-tuning improves some aspects of fidelity but does not solve the calibration problem uniformly. Mistral7B provides the strongest sentiment alignment and the highest MAUVE score, but it exceeds the real benchmark in hate prevalence. Qwen3 provides a more balanced profile across the evaluated dimensions: it improves semantic alignment, moves sentiment closer to the empirical distribution, and yields the closest hate prevalence without overshooting the benchmark. This pattern suggests that synthetic audience generation should be treated as a multi-objective calibration problem, with harmfulness, affect, semantic alignment, and diversity jointly optimized and validated against real data.

Overall, the central challenge is not whether LLMs can generate plausible reactions, but whether they can reproduce the distributional, affective, and harmfulness profiles that make real audience discourse empirically distinctive.

\section{Conclusion}\label{sec:conclusion}

This paper evaluated the extent to which LLM-powered social agents can reproduce measurable properties of real audience discourse when generating reactions to online news. Using a benchmark derived from Hatemedia, we compared real and synthetic replies elicited by the same news items across three complementary dimensions: hate speech, sentiment, and semantic alignment. In doing so, we operationalized a broader validation problem for LLM-based social simulation through a concrete large-scale case study of audience reactions to Spanish online news.

Our results show that off-the-shelf models should not be assumed to be reliable proxies of real audience behavior. In the baseline condition, all models strongly underproduced hate speech relative to the real benchmark, introduced model-specific sentiment biases, and showed limited semantic alignment with human replies. Thus, fluent and contextually plausible social-media-style text did not necessarily reproduce the harmfulness, affective profile, or semantic structure observed in real reactions to the same news stimuli.

Fine-tuning improved fidelity, but the gains were uneven and model-dependent. Qwen3 offered the most balanced profile across the evaluated dimensions, improving semantic alignment, moving sentiment closer to the empirical distribution, and producing the closest hate prevalence to the real benchmark without exceeding it. Mistral7B achieved the strongest sentiment and semantic alignment, but overshot the benchmark in hate speech. By contrast, Llama8B, Mistral24B, and GPT-OSS remained poorly calibrated in at least one core dimension despite adaptation. These findings also suggest that model scale alone is not a sufficient predictor of discourse fidelity, since the smaller Mistral7B outperformed Mistral24B in sentiment and semantic alignment after fine-tuning.

The main implication is that the realism of LLM-powered social agents should not be inferred from the plausibility of individual outputs alone. Instead, it should be empirically validated against real behavioral data and assessed at the distributional level. In this sense, synthetic audience generation should be treated as a calibration problem, where harmfulness, affect, semantic alignment, and diversity must be jointly evaluated rather than optimized in isolation.

This study has limitations. Hate speech in synthetic data was automatically estimated, whereas the real benchmark relies on human annotations; future work should include manual validation of generated replies. In addition, our evaluation focuses on Spanish news reactions and on distributional similarity, rather than on causal effects in live interaction.

Future research should therefore combine automatic metrics with human assessment, AI-generated content detection, and full OSN simulation to examine how differences in synthetic discourse affect perceived credibility, diffusion patterns, moderation outcomes, and downstream interaction dynamics. More broadly, this work shows that validating LLM-based simulations of social behavior requires comparing generated outputs with the empirical structure of real human discourse, not only evaluating whether they appear plausible in isolation.

\section*{Acknowledgments}
This work has been partially funded by the University of Murcia with the FPI/0000902983 contract.



 
\bibliographystyle{IEEEtran}
\bibliography{Computer_Society_LaTeX_template/biblio}

\newpage

\begin{IEEEbiography}[{\includegraphics[width=1in,height=1.25in,clip,keepaspectratio, angle=-90]{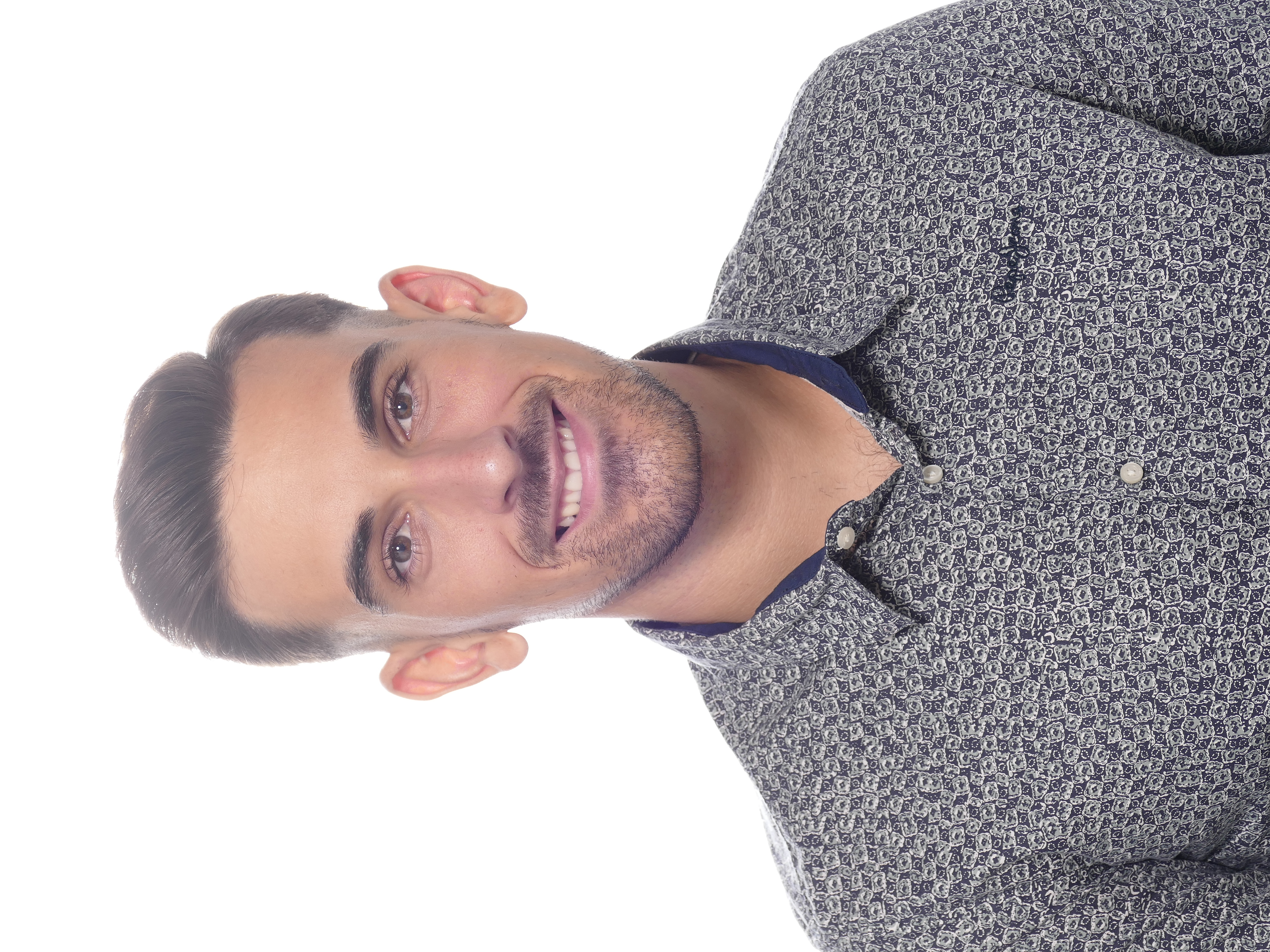}}]{Alejandro Buitrago L\'{o}pez}
is working towards a Ph.D. in Computer Science at the University of Murcia, Spain. He obtained a B.Sc. Degree with a focus on software engineering and a M.Sc. in Big Data. He is a member of the CyberDataLab at the University of Murcia, and his research interests include data science, disinformation, and cybersecurity.
\end{IEEEbiography}

\begin{IEEEbiography}[{\includegraphics[width=1in,height=1.25in,clip,keepaspectratio, angle=0]{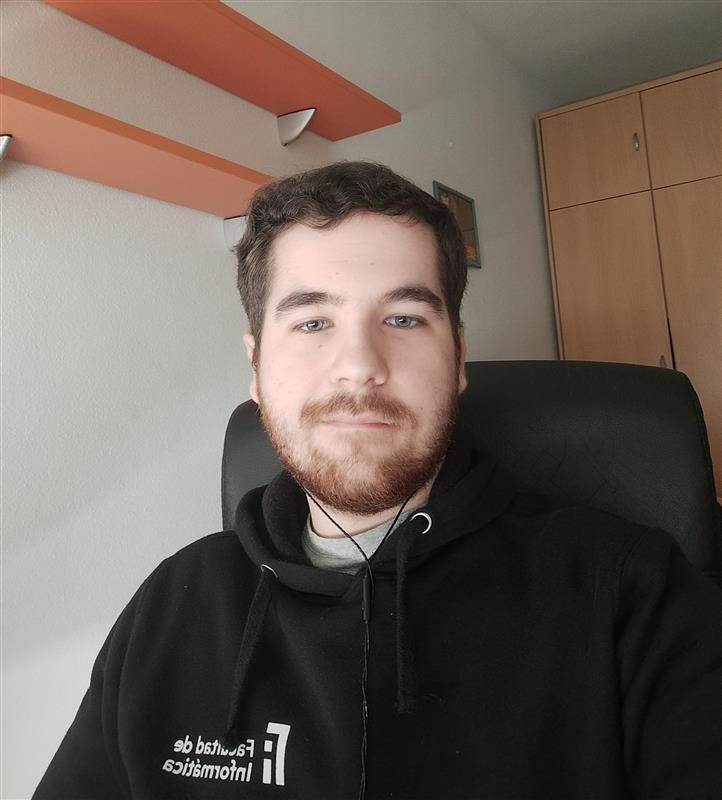}}]{Alberto Ortega Pastor} is studying a M.Sc in A.I at the University of Murcia. Degree in Computer Engineering with a focus on Computer Science. He is a member of the CyberDataLab at the University of Murcia, and his main research interests ara A.I, disinformation and machine learning.
\end{IEEEbiography}

\begin{IEEEbiography}[{\includegraphics[width=1in,height=1.25in,clip,keepaspectratio]{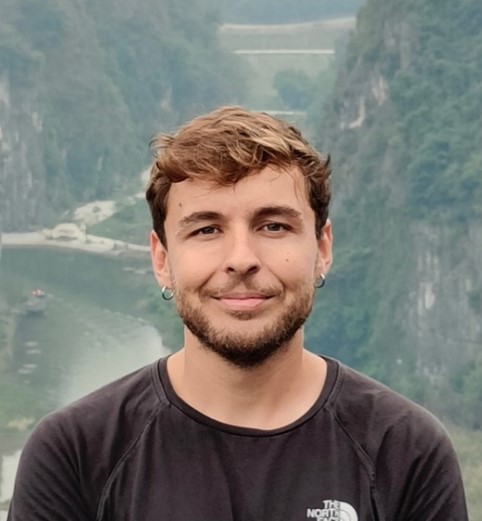}}]{Javier Pastor Galindo}  is Assistant Professor of Computer Science and Artificial Intelligence at the University of Murcia (Spain). His research focuses on combating online influence operations and mis/disinformation through artificial intelligence applications, cyber threat intelligence and simulation environments in both civilian and military contexts. Contact him at javierpg@um.es.
\end{IEEEbiography}


\begin{IEEEbiography}[{\includegraphics[width=1in,height=1.25in,clip,keepaspectratio]{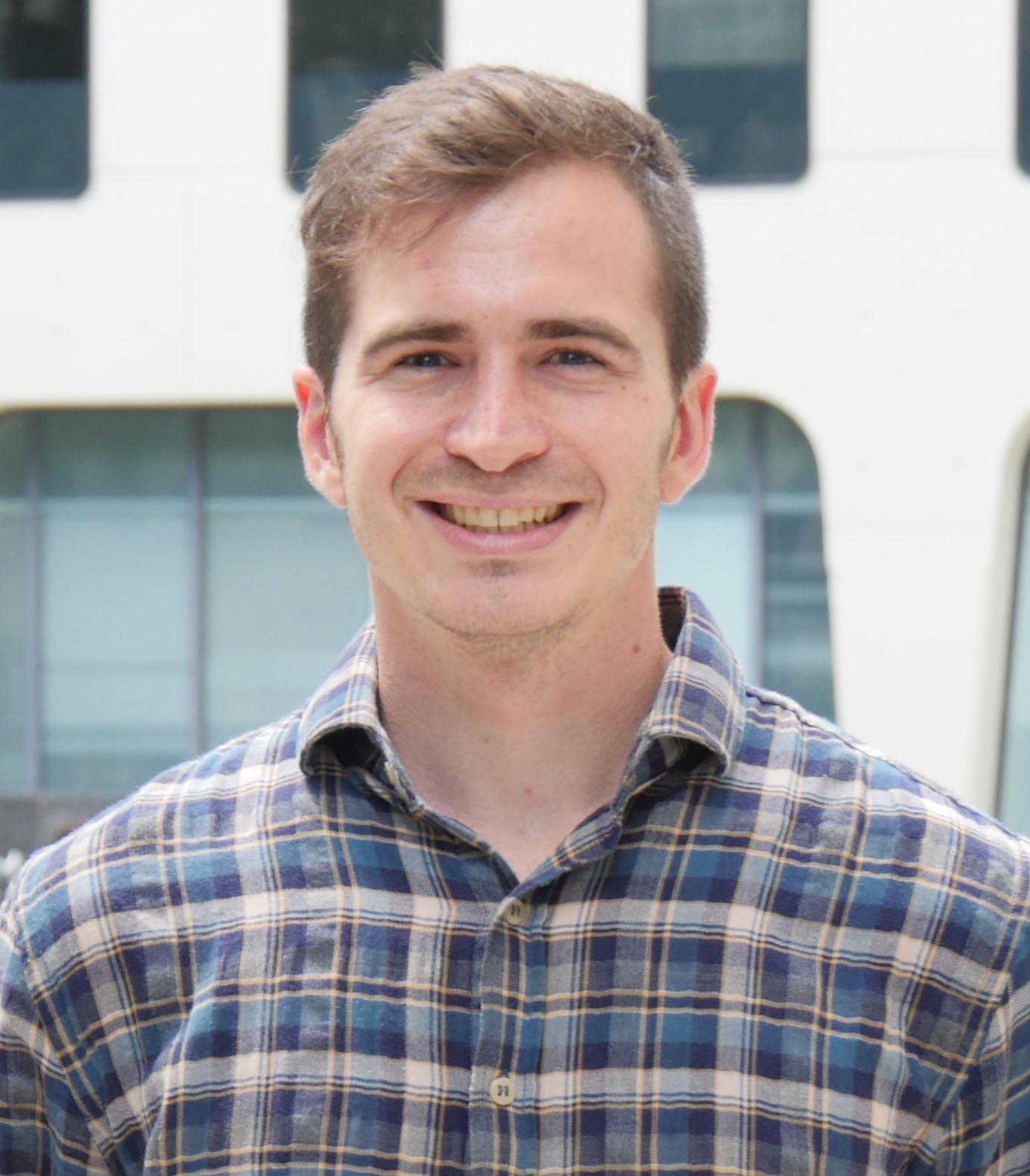}}]{Jos\'e~A.~Ruip\'erez-Valiente} (Senior Member, IEEE)
received his B.Eng. degree in telecommunications from Universidad Católica de San Antonio de Murcia in 2011 and a M.Eng. degree in telecommunications in 2013, together with his M.Sc. and Ph.D. degrees (2014 and 2017) in telematics from Universidad Carlos III of Madrid while conducting research with Institute IMDEA Networks in the area of applied data science.  He is currently an Associate Professor of Computer Science and Artificial Intelligence at the University of Murcia.
\end{IEEEbiography}

\vfill

\end{document}